\documentclass[conference]{IEEEtran}
\IEEEoverridecommandlockouts

\usepackage{amsmath}
\usepackage{amsthm}
\usepackage{newtxtext,newtxmath}   % Times, matching the template's shipped PDF
\usepackage{booktabs}
\usepackage{graphicx}
\usepackage{url}
\usepackage{cite}
\usepackage[normalem]{ulem}

\newtheorem{theorem}{Theorem}
\newtheorem{proposition}{Proposition}

\newcommand{\Slam}{S_{\lambda}}
\newcommand{\CI}[2]{[#1,\,#2]}

\begin{document}

\title{Dual Stress: Runtime Safety Monitoring for Safety-Constrained MPC Navigation
\thanks{This work was supported by the NYUAD Center for AI and Robotics (CAIR), funded by Tamkeen under the NYUAD Research Institute Award CG010.}}

\author{
\begin{tabular}{@{}p{0.47\textwidth}@{\hspace{0.04\textwidth}}p{0.47\textwidth}@{}}
\centering
{\normalsize Jamil Chahine*}\\
\textit{Department of Mechanical Engineering}\\
\textit{American University of Beirut (AUB)}\\
Beirut 1107 2020, Lebanon\\
jlc03@mail.aub.edu\\
*Corresponding author
&
\centering
{\normalsize Wenqi Cai}\\
\textit{Robotics and Intelligent Systems Control Laboratory}\\
\textit{New York University Abu Dhabi (NYUAD)}\\
Abu Dhabi, UAE\\
wenqi.cai@nyu.edu\\
\vphantom{X}
\tabularnewline[1.2em]
\centering
{\normalsize John Abanes}\\
\textit{Department of Electrical and Computer Engineering}\\
\textit{New York University}\\
New York, NY, USA\\
john.abanes@nyu.edu
&
\centering
{\normalsize Anthony Tzes}\\
\textit{Center for Artificial Intelligence and Robotics /}\\
\textit{Robotics and Intelligent Systems Control Laboratory}\\
\textit{New York University Abu Dhabi (NYUAD)}\\
Abu Dhabi, UAE\\
anthony.tzes@nyu.edu
\end{tabular}
}

\maketitle

\begin{abstract}
Runtime hazard monitors for autonomous navigation are conventionally built from geometric quantities: predicted clearance, time to collision, and required deceleration. A model-predictive controller that enforces safety through explicit constraints computes, as a by-product of every control step, a second information channel that such monitors ignore: the Karush--Kuhn--Tucker multipliers of its constrained optimization, which measure the marginal control effort spent to maintain safety against each obstacle. This paper evaluates whether a horizon-weighted sum of those multipliers, a \emph{dual stress} signal, provides a hazard monitor complementary to the geometric warnings the same state already supports. We compare it against a battery of fifteen geometric detectors tuned to a matched false-alarm budget, on preregistered held-out crossing scenarios driven through a physics simulator. The stress alarm actionably flags $4.7$ times as many collisions missed by the entire geometric battery as the geometric battery flags in return ($85$ versus $18$); combined, the two channels warn of three-quarters of the collisions for which braking remained feasible, against under half for the geometric battery alone.
\end{abstract}


\section{Introduction}

Model-predictive control with explicit safety constraints trades progress against safety by
solving one constrained program per control step; barrier-function formulations (CBF--MPC) are a
common instance \cite{ames2017cbf,zeng2021mpccbf}. Nearly every use of such a controller reads only its \emph{primal} output: the planned trajectory and the applied control. This paper asks whether
the \emph{dual} output is independently useful as a runtime hazard monitor.

The underlying principle is standard. At an optimum, a constraint's multiplier is the sensitivity of the
optimal cost to relaxing that constraint \cite[Sec.~5.6]{boyd2004convex}. For a safety constraint, that
sensitivity is the marginal control effort the optimizer is spending to maintain safety against a
specific obstacle. A large multiplier indicates that the optimizer is working hard against a
particular threat; we write $\Slam$ for the horizon-weighted sum of these
multipliers. Crucially, this is a statement about the \emph{difficulty of the escape}, not
about proximity: the same vehicle at the same distance produces no stress when stationary and substantial
stress when closing rapidly.

That distinction motivates the signal and also fixes the appropriate baseline. A safety-constrained
MPC already computes everything a distance monitor needs, so the comparison that matters is not
against a single time-to-collision threshold but against a complete battery of
predicted-clearance detectors built from the same state. And because the multiplier measures the
\emph{optimizer's} difficulty, the most informative test environment is one where the optimizer's
model departs from the plant in the ways real actuation does: tire slip, servo lag, saturating
drive torque. We therefore run the entire measurement with the controller driving a
physics-simulated plant, and ask:

\begin{quote}
\emph{On scenarios that collide, does a $\Slam$-based alarm flag cases that a strong distance-based
alarm misses in time? How many such cases arise, are the alarms causally usable, and does the
effect survive perturbation?}
\end{quote}

Our contributions are as follows.
We define dual stress for a discrete-time safety-constrained MPC and read it
from the multipliers the controller already computes at its deployed barrier
gain (Section~\ref{sec:method}). We then measure it, preregistered, against a
15-detector geometric ladder at a matched false-alarm rate on $2{,}000$ fresh
held-out scenarios driven through a physics engine
(Sections~\ref{sec:monitor}--\ref{sec:results}), execute the physical
counterfactual of braking at the alarm (Section~\ref{sec:braking}), repeat the
measurement under eleven preregistered perturbations with thresholds frozen
throughout (Section~\ref{sec:robust}), and isolate the mechanism and its failure
boundary in controlled scripted scenes (Section~\ref{sec:roads}).

\section{Related Work}

\paragraph{CBFs and CBF--MPC} Safety-critical quadratic programs built on control barrier functions \cite{ames2017cbf} were lifted to a receding-horizon setting with discrete-time barrier constraints by Zeng et al. \cite{zeng2021mpccbf}, which is the controller family we use. Recent work has extended this line toward real-time autonomous driving and stronger MPC--CBF connections: Allamaa et al. \cite{allamaa2024real} demonstrated a real-time NMPC--CBF formulation for urban autonomous driving, while Huang et al. \cite{huang2025predictive} showed that the value function of a safe MPC formulation can itself serve as a predictive control barrier function. These works strengthen the controller and its safety certificate; our focus is orthogonal: we leave the deployed controller unchanged and test whether the dual multipliers already produced by its safety constraints provide a complementary runtime hazard signal.

\paragraph{Surrogate safety measures} Time-to-collision, post-encroachment time, and deceleration-based indicators such as the brake threat number and %DRAC
{Deceleration Rate to Avoid a Crash (DRAC)} are the established surrogate safety measures \cite{hillenbrand2006btn,johnsson2018ssm,westhofen2023criticality}. The reviews are explicit that no single indicator is universal and that indicators are combined precisely because each captures only a part of the picture \cite{johnsson2018ssm}. That is why our baseline is a \emph{ladder} of predicted-clearance detectors plus rate, time-to-collision, and DRAC detectors, rather than any one indicator.

\paragraph{Preventability} 
%The counterfactual deadline we use is grounded in the inevitable collision state construct \cite{fraichard2004ics,bouraine2012ics} and in the responsibility-sensitive-safety notion of preventability under a worst-case evasive maneuver \cite{shalevshwartz2017rss}. Because our plant is a physics engine, preventability is not merely computed but \emph{executed}: the braking counterfactual of Section~\ref{sec:braking} re-runs the physics with the intervention applied.
{
The counterfactual deadline is based on two established safety concepts: inevitable collision states \cite{fraichard2004ics,bouraine2012ics} and the responsibility-sensitive-safety definition of whether a collision can still be prevented using a worst-case evasive maneuver \cite{shalevshwartz2017rss}. Since our plant is simulated with a physics engine, we not only calculate whether a collision is preventable; we test it directly by re-running the simulation with the braking intervention described in Section~\ref{sec:braking}.
}

\paragraph{Evaluation} We follow preregistration practice with frozen hypotheses, splits, and
pass/fail bars \cite{prereg2021}, and interval estimation for rare-positive proportions
\cite{brown2001interval,saito2015prroc}. 
%The strong-baseline literature shows claimed advantages shrink as baselines are tuned \cite{rendle2019progress,melis2018lstm}; our response is a baseline that is a complete ladder from the outset, tuned on the same budget as the candidate signal. 
{
Prior work shows that apparent improvements often become smaller when stronger, carefully tuned baselines are used \cite{rendle2019progress,melis2018lstm}. We therefore compare the candidate signal against a comprehensive baseline ladder from the outset, tuned under the same false-alarm budget from the start.
}
The plant model and parameters follow the published F1TENTH vehicle description \cite{trumpp2023residual}, simulated in NVIDIA Isaac Sim \cite{nvidia_isaac} with the PhysX Vehicle SDK \cite{nvidia_physx}.

\section{Dual Stress}
\label{sec:method}

\subsection{Controller}

The controller plans with a kinematic bicycle: state $x=[p_x,p_y,\psi,v]^\top$, control
$u=[a,\delta]^\top$, one forward-Euler step per tick $\Delta t$:
\begin{equation}
    \begin{aligned}
        p_x^{+}&=p_x+v\cos\psi\,\Delta t, &\quad p_y^{+}&=p_y+v\sin\psi\,\Delta t,\\
        \psi^{+}&=\psi+\tfrac{v}{L}\tan\delta\,\Delta t, &\quad v^{+}&=\max(0,v+a\,\Delta t),
    \end{aligned}
    \label{eq:euler}
\end{equation}
with wheelbase $L=0.3302$\,m. Each obstacle $j$ defines a barrier
$h_j(x)=(p_x-o_{x,j})^2+(p_y-o_{y,j})^2-r^2$ with keep-out radius $r=0.6$\,m; these are the elements
of the barrier set $\mathcal B$.
% src: campaign config.yaml ego.wheelbase, controller.safe_radius

Over a horizon of $N$ steps, the controller solves
\begin{equation}
    \begin{aligned}
        \min_{X,U,S}\ &\sum_{k=0}^{N-1}\Big[w_v(v_{k+1}-v_{\mathrm{ref}})^2+w_y(p_{y,k+1}-y_{\mathrm{ref}})^2\\
        &\quad +w_\psi\psi_{k+1}^2+w_a a_k^2+w_\delta\delta_k^2+\rho_s\!\!\sum_{i\in\mathcal B}\!s_{i,k}^2\Big]\\
        \text{s.t.}\ &X_{:,k+1}=f(X_{:,k},U_{:,k}),\quad X_{:,0}=x_0,\\
        &(1-\gamma)h_{i,k}-h_{i,k+1}\le s_{i,k},\quad s_{i,k}\ge0,\ \forall i\in\mathcal B,
    \end{aligned}
    \label{eq:nlp}
\end{equation}
subject to actuation and state bounds, solved by {Interior Point Optimizer} (IPOPT). The discrete-time barrier constraint with
gain $\gamma\in(0,1]$ requires $h$ to decay no faster than a factor $(1-\gamma)$ per step, relaxed
by a slack penalized at $\rho_s=10^{3}$. The applied control is $U_{:,0}$. The controller acts at
$\gamma=0.35$ throughout, and the multipliers below are read from that deployed controller.
% src: campaign config.yaml controller.gamma_act, controller.slack_weight

Because the road boundary enters as hard box constraints on $p_y$ and not as barrier constraints,
$\mathcal B$ contains obstacles only, and every multiplier in \eqref{eq:stress} is therefore
associated with a vehicle rather than a wall. Had the walls been soft barriers, $\Slam$ would mix obstacle threat with
lane-keeping pressure, and the monitor below would be measuring two things at once.

\subsection{Plant}
\label{sec:plant}

\begin{figure}[t]
\centering
\includegraphics[width=0.92\columnwidth]{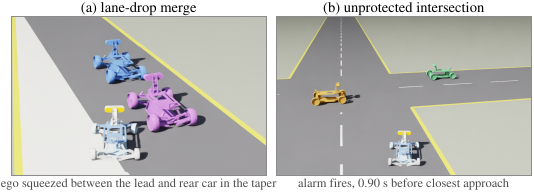}
\caption{Two scripted road scenes on the physics plant: (a)~a lane-drop merge and (b)~an
unprotected intersection, each as the conflict develops. {The white car represents the ego vehicle.}}
\label{fig:isaacshots}
\end{figure}

The controller plans with the kinematic model of \eqref{eq:euler}, but acts on a
higher-fidelity plant. The simulation runs in NVIDIA Isaac Sim \cite{nvidia_isaac}, and the ego is
a vehicle of its PhysX Vehicle SDK \cite{nvidia_physx}, with suspension and tire forces resolved by
the engine at raycast wheel contacts, parameterized from the published F1TENTH platform description
\cite{trumpp2023residual}:
mass $3.47$\,kg, wheelbase $0.3302$\,m, length $0.51$\,m, friction $0.8$. Steering is a servo: a
first-order lag with time constant $0.08$\,s under a rate limit of $3.2$\,rad/s, and drive/brake
torques are scaled by two calibration constants fitted once on separate calibration maneuvers and
frozen before any campaign data were collected. The physics steps at $10$\,ms, five substeps per
$50$\,ms control tick. The controller therefore plans with a model that omits tire slip, servo lag, and torque saturation. Traffic vehicles follow their scripted
constant-velocity trajectories exactly and collide as rigid boxes of the drawn footprint
(Fig.~\ref{fig:isaacshots}).
% src: campaign config.yaml vehicle.*, physics.*

\subsection{The signal}

At the solution, each barrier constraint carries a multiplier $\lambda_{i,k}\ge0$, read directly from
the solver per named constraint. Its meaning is standard \cite[Sec.~5.6]{boyd2004convex}:

\begin{theorem}[Value sensitivity]
\label{thm:sens}
For \eqref{eq:nlp}, the sensitivity of the optimal cost $V^\star$ to relaxing barrier constraint $i$ is
$\partial V^\star/\partial b_i=-\lambda^\star_i$. The multiplier is the marginal control effort required to enforce barrier $i$.
\end{theorem}

The following one-dimensional braking case provides an exact interpretation of the multiplier.

\begin{proposition}[1-D braking]
\label{prop:brake}
For head-on braking with nominal acceleration $a_{\mathrm{nom}}$, gap $d$, speed $v$, gain $\gamma$
and minimum distance $d_{\min}$, the barrier-filter problem
$\min_a\tfrac12(a-a_{\mathrm{nom}})^2$ s.t.\ $a\le a_{\mathrm{safe}}$ has
\begin{equation}
a_{\mathrm{safe}}=\tfrac{2}{\Delta t^2}\big(\gamma(d-d_{\min})-v\Delta t\big),\,\,
\lambda^\star=\max\!\big(0,a_{\mathrm{nom}}-a_{\mathrm{safe}}\big).
\label{eq:prop2}
\end{equation}
\end{proposition}

Equation~\eqref{eq:prop2} shows why the signal measures difficulty rather than proximity:
$\lambda^\star$ is the gap between the deceleration the task wants and the deceleration safety
allows, and it grows with speed at fixed distance. A unit test asserts that the solver's extracted
$\lambda$ matches \eqref{eq:prop2} to $10^{-4}$, validating the multiplier extraction on which
every downstream measurement depends.

Per-step dual stress sums the multipliers over the horizon in physical barrier-relaxation units,
\begin{equation}
\Slam(t)=\sum_{i,k}\rho_{\mathrm{hz}}^{\,k}\,\lambda_{i,k}\,\Delta b,
\qquad \Delta b=r^2-(r-\delta r)^2,
\label{eq:stress}
\end{equation}
with horizon decay $\rho_{\mathrm{hz}}=0.95$ and $\Delta b$ the physical relaxation corresponding to
shrinking the keep-out radius by $\delta r$.

The runtime alarm statistic is the baseline-subtracted stress
$\mathrm{F1}(t)=\Slam(t)-\operatorname{median}(\Slam(0{:}10))$, sliced before the first barrier
violation. When IPOPT returns an infeasible iterate, no multipliers exist for that tick; the monitor
holds the last valid value instead, a rule frozen before data collection. Nothing is added to the control
loop to obtain the signal: $\Slam$ is read from the multipliers of the single solve the controller
already performs each step, so the monitor is free at runtime.

\section{The Monitor Comparison}
\label{sec:monitor}

\subsection{A complete geometric ladder}

The baseline is everything a distance monitor can compute from the same state. Extrapolating the ego
and every obstacle at constant velocity, the predicted clearance level at horizon $k$ is
\begin{equation}
\mathrm{level}(k)=-\min_{j,\,m\le k}\big[(\hat e_x-\hat o_{x,j})^2+(\hat e_y-\hat o_{y,j})^2-r^2\big].
\label{eq:level}
\end{equation}
The ladder has 15 detectors in total: 12 level detectors at horizons $k\in\{4,6,8,12,16,22\}$ each paired with two persistence settings (an alarm must hold for 1 or 6 consecutive steps), one \emph{rate} detector, the lag-2 derivative of $\mathrm{level}(22)$, one time-to-collision detector, and one DRAC detector \cite{johnsson2018ssm}. Every threshold is tuned on the training split to maximize the true-positive rate subject to a false-positive rate $\le10\%$ on the benign stratum (scenarios whose closest approach stays above half the squared keep-out radius); the identical budget is applied to the stress alarm. That matched budget is what makes this a signal comparison rather than a threshold choice. The geometry channel alarms when \emph{any} of the 15 detectors fires.

\subsection{Deadline and actionability}

Two kinematic quantities, computed without the optimizer, decide whether a warning could have been used. The brake threat number $a_{\mathrm{req}}/a_{\max}^{\mathrm{brake}}$ crosses $1$ when braking alone can no longer avoid contact; the last tick before that crossing is the last-feasible-brake deadline $t_L$. A colliding scenario whose brake threat number crosses $1$ only after the initial tick has a well-defined deadline; we call such collisions \emph{brakeable}, and only they can be warned of at all. An alarm is \emph{actionable} if it fires at least $0.1$\,s before $t_L$. Each brakeable collision is then labelled by which channel fires actionably: \textbf{stress-only}, \textbf{distance-only}, \textbf{both}, or \textbf{neither}. Proportions carry Wilson score intervals \cite{brown2001interval}. Section~\ref{sec:braking} adds the stronger, physics-executed test: whether braking at the alarm actually prevents the collision.

\subsection{Design and operating point}

We randomly generate six-car crossing-traffic scenarios: each scenario samples crossing angles,
speeds, and times-to-conflict for six cars, several of which pose genuine conflicts while the rest serve as distractors. The three splits are disjoint by construction and were frozen before collection: a
pilot of $100$ scenarios used only for operational checks and excluded from inference; a training
split of $500$ on which all $16$ thresholds were tuned; and a test split of $2{,}000$ opened only
after every threshold was frozen. The preregistered acceptance criterion is a stress-only rate
whose $95\%$ lower confidence bound exceeds $0.5\%$. The controller is fixed throughout: horizon
$N=22$, $\Delta t=0.05$\,s, keep-out $r=0.6$\,m, nominal speed $3$\,m/s, hard road bounds at
$\pm1.2$\,m.
% src: campaign config.yaml world.*, controller.horizon; statistical_verdict.json n_train/n_test

Three properties of the harness were verified before the campaign and underpin every result {in Section~\ref{sec:results}}: rollouts are bit-exact deterministic across processes; the recorded stress round-trips to the solver's raw multipliers with zero error; and rendering the scene does not change the physics (viewer and headless runs are bit-identical).

\section{Results}
\label{sec:results}

\subsection{Coverage on fresh held-out data}
\label{sec:headline}

\begin{figure}[t]
\centering
\includegraphics[width=0.92\columnwidth]{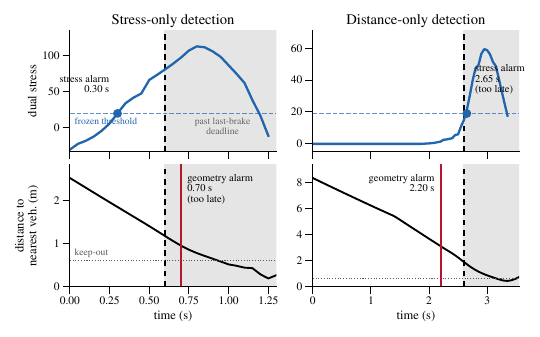}
\caption{A stress-only detection (left) and a distance-only detection (right) on the physics
plant. Top: dual stress and its frozen threshold; the dot marks the alarm.
Bottom: distance to the nearest vehicle, with the earliest geometric-ladder alarm as a solid
vertical line. In both, the dashed line is the last-brake deadline and the shaded region is past
it. The winning channel alarms before the deadline and the losing channel after it, so the two
scenarios are mirror images of each other.}
\label{fig:cases}
\end{figure}

Of the $2{,}000$ test scenarios, $316$ end in a collision, and $309$ of these are \emph{brakeable}: braking remained feasible past the first tick, so a warning could in principle have been acted upon. The remaining scenarios either do not collide or leave no feasible braking window, and are not classifiable. Each of the $309$ brakeable collisions falls into exactly one of four classes according to which channel warns in time---stress only ($85$), distance only ($18$), both ($128$), or neither ($78$)---and these counts partition the $309$. {Breakdown is shown in Table~\ref{tab:headline}.}

The preregistered primary result is the stress-only rate, $85$ of the full $2{,}000$ scenarios, or
$4.25\%$ $\CI{3.45}{5.23}$:
% src: statistical_verdict.json:stress_only_rate, stress_only_wilson_95
collisions the stress alarm flagged in time while the entire 15-detector ladder did not. We express this as a fraction of all $2{,}000$ scenarios, not of the $309$ collisions, because that matches the
preregistered criterion, which bounds how often such a collision arises in the scenario population as a whole. The lower confidence bound, $3.45\%$, exceeds the preregistered $0.5\%$ criterion by roughly sevenfold. The converse,
distance-only, rate is $0.90\%$ $\CI{0.57}{1.42}$ ($18$ scenarios). Of the $78$ collisions that
neither channel warned of in time, every one nevertheless produced an alarm, with a median best
alarm only $0.05$\,s past the deadline: the conflict developed too quickly for either channel to
provide a timely warning.
% src: statistical_verdict.json:counts; caught_rows.json neither-class best-lead median

Expressed as coverage of the $309$ brakeable collisions, the single stress detector actionably warns of
$213$ ($69\%$ $\CI{63.6}{73.8}$), the 15-detector ladder of $146$ ($47\%$ $\CI{41.8}{52.8}$), and
their union of $231$ ($75\%$).
% src: derived from statistical_verdict.json:counts; Wilson recomputed by check_numbers.py
Both channels are individually tuned to the same $10\%$ benign false-alarm budget. A union
monitor has not been separately tuned to a matched budget on this plant, so the union figure is
descriptive rather than a deployment claim.
% \begin{table}[t]
% \centering
% \caption{Detection outcomes on the preregistered test split of $2{,}000$ held-out scenarios. Thresholds frozen on the disjoint training split. Rates are fractions of all $2{,}000$ scenarios; coverage and rescue use their own denominators. Intervals are at the $95\%$ level.}
% \label{tab:headline}
% \begin{tabular}{lcc}
%     \toprule
%      & count & \% of $2{,}000$ \\
%     \midrule
%     \multicolumn{3}{l}{\emph{Detection of the $309$ brakeable collisions}} \\
%     stress only        & $85$  & $4.25\%$ $\CI{3.45}{5.23}$ \\
%     distance only      & $18$  & $0.90\%$ $\CI{0.57}{1.42}$ \\
%     both channels      & $128$ & $6.40\%$ \\
%     neither in time    & $78$  & $3.90\%$ \\
%     \addlinespace
%     \multicolumn{3}{l}{\emph{Coverage of the $309$ brakeable collisions}} \\
%     stress channel     & \multicolumn{2}{c}{$213/309 = 69\%$} \\
%     geometric ladder   & \multicolumn{2}{c}{$146/309 = 47\%$} \\
%     \addlinespace
%     \multicolumn{3}{l}{\emph{Braking rescue of the $85$ stress-only cases}} \\
%     collisions avoided & \multicolumn{2}{c}{$82/85 = 96.5\%$ $\CI{90.1}{98.8}$} \\
%     \bottomrule
%     \end{tabular}
% \end{table}
\begin{table}[t]
    \centering
    \caption{Detection outcomes on the preregistered test split of $2{,}000$ held-out scenarios.}
    \label{tab:headline}
    
    \small
    \renewcommand{\arraystretch}{1.08}
    
    \begin{tabular*}{\linewidth}{@{\extracolsep{\fill}}lcc@{}}
        \toprule
        Outcome
        & Count
        & Rate \\
        \midrule
        
        \multicolumn{3}{@{}l}{\textit{Detection outcomes: 309 brakeable collisions}} \\
        \quad Stress only
        & $85$
        & $4.25\%$ $\CI{3.45}{5.23}$ \\
        \quad Distance only
        & $18$
        & $0.90\%$ $\CI{0.57}{1.42}$ \\
        \quad Both channels
        & $128$
        & $6.40\%$ \\
        \quad Neither in time
        & $78$
        & $3.90\%$ \\
        
        \midrule
        \multicolumn{3}{@{}l}{\textit{Coverage: 309 brakeable collisions}} \\
        \quad Stress channel
        & $213$
        & $69\%$ \\
        \quad Geometric ladder
        & $146$
        & $47\%$ \\
        
        \midrule
        \multicolumn{3}{@{}l}{\textit{Braking rescue: 85 stress-only cases}} \\
        \quad Collisions avoided
        & $82$
        & $96.5\%$ $\CI{90.1}{98.8}$ \\
        
        \bottomrule
    \end{tabular*}

    \vspace{1pt}

    \begin{minipage}{\linewidth}
    \footnotesize
    \raggedright
    Thresholds frozen on the disjoint training split. Rates are fractions of all $2{,}000$ scenarios; coverage and rescue use their own denominators. Intervals are at the $95\%$ level.
    \end{minipage}
\end{table}
Figure~\ref{fig:cases} shows a representative case of each class side by side,
making the complementarity concrete: in each scenario, one channel gives the
only timely warning and the other reacts too late.
% src: test scenarios 896 (stress-only) and 259 (distance-only); alarm/deadline times in the figure
The same ordering holds across the actionability margin: stress covers more brakeable collisions
than the ladder at \emph{every} required lead up to about $0.45$\,s, the two cross near $0.5$\,s,
and the ladder retains a long-lead tail on threats visible well in advance.
% src: caught_rows.json stress_only median s_lead; margin curve crossing, plot_margin_exemplar_paper.py

\begin{figure}[t]
\centering
\includegraphics[width=0.92\columnwidth]{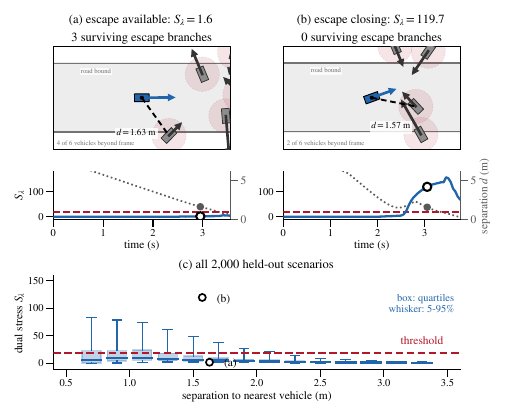}
\caption{Stress is not a function of separation. (a)~and~(b) are two held-out scenarios at nearly identical separation: three escape branches survive in (a), none in (b), and stress is roughly
$70\times$ higher. (c)~All $2{,}000$ held-out scenarios binned by separation.}
\label{fig:decoupling}
\end{figure}
% src: plots/decoupling_facts_c.json:stress_ratio (74.27 -> "70x"), calm/stressed survivors 3/0

The contrast in the two cases holds across the whole test set. Stress-only alarms fire at close
range with high peak stress (median crossing distance $1.84$\,m, median peak stress $111$), while
distance-only alarms fire farther out on scenarios with lower peak stress (median $2.51$\,m, peak
$73$). The optimizer's difficulty concentrates precisely where extrapolated geometry is weakest: close-range conflicts, when constant-velocity prediction has little time to resolve before contact.
% src: caught_rows.json per-class medians (s_dist, g_dist, peak_stress)

That the two channels can disagree at all is a property of the signal, not of the threshold.
Separation explains little of the variance in stress across the test set
($r=-0.39$), and the disagreement survives at matched geometry
(Fig.~\ref{fig:decoupling}). Binned by separation, the stress distribution
straddles the alarm threshold at every distance. Proximity therefore does not determine constraint pressure, which is
why a stress-only class exists to be populated at all.
% src: plots/decoupling_facts_c.json:pearson_r_separation_stress, stress_ratio

\subsection{The alarms are causally usable}
\label{sec:braking}

An actionability label derived from a deadline is still only a label. Because the plant is a physics
engine, we can execute the counterfactual: each of the $85$ stress-only scenarios was re-run with a
full brake commanded from the stress-alarm tick onward, physics otherwise identical. The collision
is avoided in $82$ of $85$ cases: a rescue rate of $96.5\%$ $\CI{90.1}{98.8}$.
% src: braking_verdict.json:rescued, n_candidates
These alarms are not post-hoc correlates; acting on them prevents the crash. One caveat is stated
plainly: a matched control arm (identical braking triggered at times when no alarm fired) has not
yet been run, so the formal causal gate on the monitor remains open by design, and the three
unrescued cases have not been individually diagnosed.
% src: braking_verdict.json:monitor_gate_passed, note

\subsection{Robustness: eleven perturbations, thresholds never retuned}
\label{sec:robust}

The full test split was re-run under eleven preregistered conditions, each a single controlled
deviation from the nominal setup, with every threshold held at its nominal-training value. Five
perturb the observations the monitor receives: a sensor latency of $50$\,ms and of $100$\,ms, which
delay every measurement by that interval; additive Gaussian noise on the measured ego and obstacle
states; a $5\%$ per-step probability that a measurement is dropped; and a \emph{combined} condition
that applies the $100$\,ms latency, the noise, and the dropout together. Six perturb the plant
itself, one factor at a time: tire friction scaled by $0.75$ and by $1.25$, vehicle mass and yaw
inertia by $0.8$ and $1.2$, and actuation authority---the drive and brake torque limits---by $0.8$
and $1.2$. Table~\ref{tab:robust} reports the outcome. The stress-only rate remains above the $0.5\%$ criterion in every condition.
\begin{table}[t]
    \centering
    \caption{Robustness under frozen thresholds.}
    \label{tab:robust}
    
    \small
    \setlength{\tabcolsep}{3.5pt}
    \renewcommand{\arraystretch}{1.05}
    
    \begin{tabular}{@{}lrrrr@{}}
        \toprule
        Condition
        & Brakeable
        & Stress-only
        & All (\%)
        & Brakeable (\%) \\
        \midrule
        
        \multicolumn{5}{@{}l}{\textit{Nominal}} \\
        \quad Nominal
        & 309 & 85 & 4.25 & 27.5 \\
        
        \midrule
        \multicolumn{5}{@{}l}{\textit{Observation perturbations}} \\
        \quad Latency, $50$\,ms
        & 578 & 147 & 7.35 & 25.4 \\
        \quad Latency, $100$\,ms
        & 969 & 192 & 9.60 & 19.8 \\
        \quad Noise
        & 289 & 74 & 3.70 & 25.6 \\
        \quad Dropout, $5\%$
        & 334 & 88 & 4.40 & 26.3 \\
        \quad Combined
        & 1066 & 181 & 9.05 & 17.0 \\
        
        \midrule
        \multicolumn{5}{@{}l}{\textit{Plant perturbations}} \\
        \quad Friction $\times 0.75$
        & 402 & 94 & 4.70 & 23.4 \\
        \quad Friction $\times 1.25$
        & 317 & 85 & 4.25 & 26.8 \\
        \quad Mass $\times 0.8$
        & 305 & 84 & 4.20 & 27.5 \\
        \quad Mass $\times 1.2$
        & 306 & 83 & 4.15 & 27.1 \\
        \quad Authority $\times 0.8$
        & 346 & 92 & 4.60 & 26.6 \\
        \quad Authority $\times 1.2$
        & 309 & 85 & 4.25 & 27.5 \\
        
        \bottomrule
    \end{tabular}

    \vspace{1pt}
    
    \begin{minipage}{0.98\columnwidth}
        \footnotesize
        \raggedright
        Thresholds are frozen at nominal. The five observation conditions comprise $50$\,ms latency, $100$\,ms latency, noise, dropout, and combined ($100$\,ms latency, noise, and dropout); the last six scale a single plant factor. ``Brakeable(\%)'' is the share of that condition's brakeable collisions caught only by stress, the scale-invariant quantity.
    \end{minipage}
\end{table}
Two observations follow. First, the share of scenarios rises under degradation only because
degradation increases the number of collisions; the share of \emph{brakeable} collisions is the
invariant quantity, and it remains at $23.4$--$27.5\%$ across every plant perturbation and every
mild observation condition. Plant uncertainty in friction, mass, and actuation authority changes
the result only marginally ($4.15$--$4.70\%$ of scenarios). This is the most important result in the table: the signal is insensitive to model--plant mismatch in the vehicle, which is the
regime the mechanism is intended to address.
Second, the two heavy sensor corruptions are the only conditions that erode the stress share
($19.8\%$ and $17.0\%$ of brakeable), and in exactly those conditions the ladder's unique share
rises sharply (to $169$ and $308$ scenarios respectively): under degraded observations the two
channels cease to be redundant and become strongly complementary. Sensing corruption remains the
signal's principal weakness, and its compensation by the geometric channel is visible in the same
table.
% src: robustness/latency_100ms/verdict.json:counts.geometry_only, robustness/combined/verdict.json:counts.geometry_only

\subsection{Mechanism and boundary: scripted road scenes}
\label{sec:roads}
\begin{figure*}[t]
\centering
\includegraphics[width=0.84\textwidth]{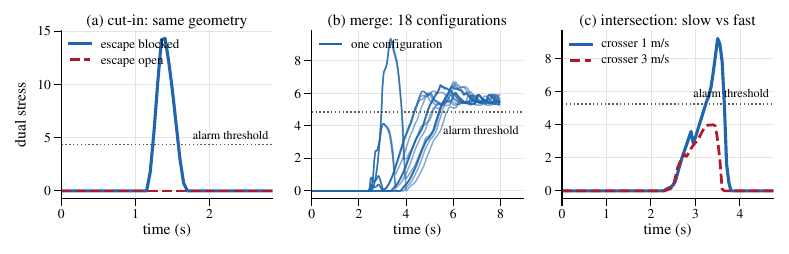}
\caption{Scripted road scenes on the physics plant: dual stress over time against each family's
frozen alarm threshold (dotted). (a)~Cut-in, one pair with identical ego--intruder geometry: the
escape-blocked run crosses the threshold while the escape-open run stays silent. (b)~Merge, all
$18$ configurations (a $3\times3\times2$ grid over the two cars' start positions and the closing
speed): every one crosses the threshold, a broad regime rather than a knife-edge.
(c)~Intersection: a slow crosser ($1$\,m/s) crosses the threshold, while a fast crosser
($3$\,m/s) with the same conflict but half the observation window never does.}
\label{fig:roadscenes}
\end{figure*}

Statistics on random traffic establish how often the effect occurs; controlled scenes establish
why. Three scripted families, a highway cut-in, a lane-drop merge, and an urban intersection,
were run on the same plant with per-family thresholds frozen from family-specific benign pools
under the same $10\%$ rule (Fig.~\ref{fig:roadscenes}).

\paragraph{Cut-in: the mechanism} The central experiment is a \emph{paired} design. In each of six
pairs, the cutting-in car's trajectory relative to the ego is identical; the only difference is a
neighbouring lane that is either open or blocked by a pacing car. With the escape open, stress remains at
$0.0$ throughout and the alarm stays silent; with the escape blocked, stress reaches ${\sim}14$
and the alarm fires, in all six pairs (Fig.~\ref{fig:roadscenes}a). The geometry is identical and the verdicts are opposite:
the multiplier reflects the \emph{feasibility of the escape} rather than the proximity of the
threat, which is the controlled-scene analogue of Proposition~\ref{prop:brake}. The single road-scene collision in the campaign, a severe cut-in with the escape blocked, was
preceded by a stress alarm $2.35$\,s before impact.
% src: road/cutin_matrix_physics.json paired cells; C4 alarm lead

\paragraph{Merge: a robust regime} We sweep a grid of $18$ merge configurations---the $3\times3\times2$
combinations of the lead car's start position, the closing car's start position, and the closing
speed. Every configuration ends in a near-miss (closest approach within $\pm0.08$ barrier units of
contact) and stress reads $5.9$--$9.4$ in all of them (Fig.~\ref{fig:roadscenes}b): the merge is a
broad stress-positive regime rather than a narrow parameter region. In a closed-loop test, a yield maneuver triggered
by the frozen stress alarm executes legally and safely on the physics plant, although it does not
increase the realized clearance margin; triggering an acceleration instead drives the vehicle off
the road, indicating that the alarm supports cautious interventions rather than aggressive ones.
% src: road/merge_grid_physics.json; merge interventions M3/M5

\paragraph{Intersection: the boundary} An occluded crossing at $1.5$\,m/s draws a stress alarm
$0.9$\,s before the conflict, but as the crosser's speed is swept the stress lead shrinks and then
vanishes above approximately $2$\,m/s (Fig.~\ref{fig:roadscenes}c), while the geometric lead stays
near $4$\,s. The failure is
structural, not a threshold artifact: constraint pressure accumulates over the horizon, so the
signal needs roughly one second of observation to register a threat. These constant-velocity scenes
are also the geometric ladder's best case, so they bound the claim from both sides.
% src: road/intersection_hardness_physics.json sweep

\section{Scope and Limitations}
\label{sec:limits}

\paragraph{Role of the signal} Dual stress complements geometric monitoring rather than replacing
it. Its strength is selectivity near the last-brake deadline, where it flags close-range,
dynamically difficult conflicts that constant-velocity prediction resolves only late. Geometric
detectors retain the advantage on threats that are visible well in advance.
Because constraint pressure builds over the horizon, the signal is most informative once about a
second of observation is available (Section~\ref{sec:roads}), and some conflicts develop too
quickly for either channel to warn in time. We study it as a runtime monitor read from the existing
solution; whether incorporating it into the control law is beneficial is a separate question we
leave to future work.

\paragraph{Scope} Every statistical result concerns six-car crossing traffic on a physics-simulated
1/10-scale plant with scripted constant-velocity obstacles; the road scenes are scripted
single-conflict scenes on the same plant. Because the obstacles move at constant velocity, the
geometric ladder's constant-velocity extrapolation is exact, which gives the baseline an
advantage it would not enjoy against maneuvering traffic. The results do not address reactive
traffic, full-size vehicles, or real perception, and the advantage reported here should be
expected to vary with the size of the model--plant gap.

\paragraph{Open points} Three items remain for future work. First, the braking counterfactual of
Section~\ref{sec:braking} shows that braking at the alarm prevents most collisions, but a matched
comparison---braking at times when no alarm fired---is still needed to attribute the prevention to
the alarm rather than to braking itself; three stress-only collisions that braking did not prevent
also remain to be examined. Second, a combined monitor that alarms whenever either channel fires
was not separately tuned to a matched false-alarm budget, so its coverage in Table~\ref{tab:headline}
is reported descriptively rather than as a deployment result. Third, transfer to the physical
F1TENTH platform from which the plant parameters are taken is the natural next step.

\section{Conclusion}

The dual variables of a safety-constrained model-predictive controller form a usable runtime
hazard monitor, and the signal is most valuable precisely where the controller's model departs
from the plant, as in the present case of a CBF--MPC driving a physics-simulated vehicle. On $2{,}000$
preregistered held-out crossing scenarios, a single frozen-threshold stress alarm actionably
flagged $4.25\%$ of scenarios whose collisions a complete 15-detector geometric ladder missed,
covered $69\%$ of brakeable collisions against the ladder's $47\%$, and braking on its alarms
prevented $96.5\%$ of the collisions only it detected. The result persisted across eleven
preregistered perturbations without retuning, was insensitive to plant uncertainty, and degraded
only under heavy sensor corruption, where the geometric channel's complementary contribution
grows. Paired scripted scenes connect the statistics to the mechanism: the signal measures the
feasibility of escape rather than the proximity of the threat, and it is silent when the threat
develops faster than the horizon can register. The claim is deliberately bounded: this plant, this
scale, this traffic. The alarms are late but selective; and the signal's boundary is reported
alongside its value.

\bibliographystyle{IEEEtran}
\bibliography{refs}

\end{document}